\documentclass[runningheads]{llncs}
\usepackage[T1]{fontenc}
\usepackage{graphicx}
\usepackage{booktabs}
\usepackage{multirow}
\usepackage{float}
\usepackage{amsmath}
\usepackage{amssymb}
\usepackage{listings}
\usepackage{xcolor}

\begin{document}
\title{GRACE: Boosting Video MLLMs with Grounded Action-Centric Evidence for Viewer Sentiment Prediction}
\titlerunning{GRACE for Viewer Sentiment Prediction}


\author{Ruoxuan Yang\inst{1} \and
Tieyuan Chen\inst{1} \and
Xiaofeng Huang\inst{2} \and
Haibing Yin\inst{2} \and
Jun Wang\inst{3} \and
Xiping Chen\inst{4} \and
Jun Yin\inst{5} \and
Xuesong Gao\inst{6,7} \and
Weiyao Lin\inst{1}}

\authorrunning{R. Yang et al.}

\institute{
Shanghai Jiao Tong University, China
\and
Hangzhou Dianzi University, China
\and
The 52nd Research Institute of China Electronics Technology Group Corporation, China
\and
Hangzhou Bywin Technology Co., Ltd., China
\and
Zhejiang Dahua Technology Co., Ltd., China
\and
School of Information Science and Engineering, Shandong University, China
\and
Haihe Laboratory of Information Technology Application Innovation, China
}

\maketitle              
\vspace{-1.0em}

\begin{abstract}
Viewer sentiment prediction in video advertisements aims to infer the latent affective response evoked in the audience. To bridge the gap between what is shown and what is felt, models must deduce hidden viewer emotions from explicit visual narratives, concrete character--object interactions, and visible textual cues. However, standard Multimodal Large Language Models (MLLMs) typically rely on holistic frame representations, which leave these fine-grained, affect-relevant events implicit and complicate precise emotional reasoning.
To address this, we propose a grounded action-centric evidence augmentation framework that enhances video MLLMs' clue extraction and comprehension by introducing explicit event structure and localized visual evidence. 
Our method extracts temporally ordered subject--verb--object (SVO) triplets and auxiliary visible textual cues from action-centric video descriptions, grounds subject and object entities as visual entity crops, and then enables the MLLM to perform clue-enhanced emotional reasoning based on these extracted structured clues. In this way, action triplets specify ``what happens'', while grounded visual entity crops anchor ``who or what participates in each event'' to concrete visual evidence.
Experiments on the Pitts dataset show consistent improvements over Qwen2.5-VL and Qwen3-VL baselines. Ablation studies, cross-dataset evaluation on AdsQA, and transfer experiments on an emotion-focused TVQA subset further support the effectiveness and generalization of our approach.

\end{abstract}

\section{Introduction}
\begin{figure}[t]
  \centering
  \includegraphics[width=\textwidth,trim={35 700 380 30},clip]{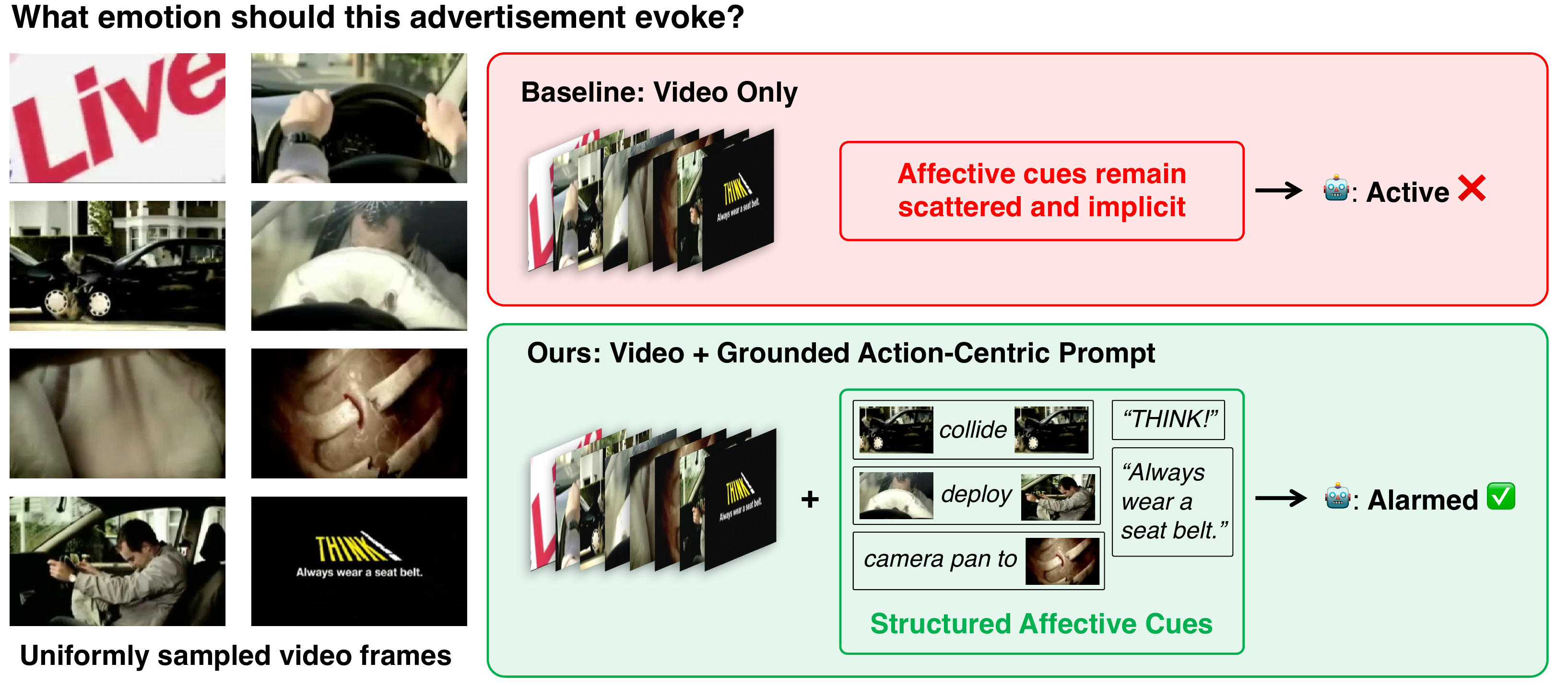}
  \caption{Illustration of sparse frame-only input and our grounded action-centric evidence augmentation. Our method makes action events and visible textual cues explicit, helping the MLLM infer the intended viewer sentiment.}
  \label{fig:intro}
\end{figure}
Viewer sentiment prediction in video advertisements is crucial for evaluating advertising effectiveness, because viewers' affective responses influence attitudes, purchase intention, and affective retention~\cite{poels2006capture,kujur2018emotions,hamelin2017emotion}. 
Unlike directly observable character emotions, viewer sentiment is a latent affective response and cannot be directly observed from the screen. 
Existing advertisement emotion models and video understanding pipelines commonly rely on global video-level features or sparsely sampled frame inputs~\cite{hussain2017automatic,antonov2024decoding,li2024mvbenchcomprehensivemultimodalvideo}. 
While compact, such holistic representations leave event-level narrative evidence implicit, requiring the model to recover affect-relevant events.
A common alternative is to rely on facial expressions or body gestures, which primarily reflect the emotions of on-screen characters and indirectly shape viewers' affective responses~\cite{fang2025emoe}.

However, these character-level emotional cues are not reliable indicators of viewer sentiment, as illustrated by the Kuleshov effect, where identical facial expressions can lead to different emotional interpretations depending on the surrounding scenes~\cite{mobbs2006kuleshov}. In video advertisements, affective cues may lie in narrative events, character--object interactions, or visible textual cues, rather than in the global appearance captured by sparsely sampled frames. These observations motivate explicit action-event organization: instead of leaving affect-relevant events implicit in sparse visual inputs, we represent video context through temporally ordered subject--verb--object triplets and auxiliary visible textual cues.

Even with explicit event organization, a purely textual representation remains limited. Subject--verb--object triplets provide a compact symbolic abstraction of actions, but entity names alone may discard the concrete visual appearance and local context of the participating entities that shape affective interpretation in advertisements. This limitation echoes the symbol grounding problem~\cite{HARNAD1990335,chen2025looking}: symbolic references should be connected to perceptual evidence rather than remaining purely textual. We therefore ground triplet subjects and objects into localized visual crops, allowing the model to reason over both symbolic event structure and concrete visual evidence.

Taken together, the context dependence of viewer sentiment and the perceptual limitation of purely symbolic triplets motivate grounded action-event reasoning for video advertisement sentiment prediction, where action-centric triplets make the narrative context explicit and visual grounding reconnects subject and object entities to localized perceptual evidence. To instantiate this formulation, we propose a grounded action-centric evidence augmentation framework. Our method first extracts temporally ordered action triplets and visible textual cues from captions generated with dense frame sampling. It then grounds subject and object entities as localized visual crops from the same sparse frames used by the MLLM input. Finally, the original sparse frames, serialized grounded action events, and visible textual cues are jointly provided to the MLLM for viewer sentiment reasoning. Figure~\ref{fig:intro} illustrates how our method makes action-level evidence explicit beyond sparse frame-only evidence.

\section{Related Work}
\subsection{Viewer Sentiment Prediction in Advertisements}
Viewers' emotional responses have long been studied in advertising research because they influence attitudes toward advertisements, purchase intention, and memory of advertised content~\cite{poels2006capture,kujur2018emotions,hamelin2017emotion}.
To support computational advertisement understanding, prior work has introduced large-scale advertisement datasets and benchmarks covering topics, sentiments, symbolic references, persuasive messages, and marketing-oriented reasoning~\cite{hussain2017automatic,long2025adsqa}.
Recent studies have also explored large-scale viewer sentiment modeling and explainable emotion detection in video advertisements~\cite{antonov2024decoding,vanneste2024detecting}.

Different from these works, which generally rely on holistic video/audio representations, viewer reaction signals, or frame-level explanations, our method explicitly organizes advertisement videos into action-event sequences with grounded visual entity evidence, making both event structure and participating entities explicit for MLLM-based viewer sentiment prediction.

\subsection{MLLMs for Video Understanding}
Recent multi-modal large language models (MLLMs) have extended language-based reasoning to visual inputs, enabling tasks such as video captioning, open-ended question answering, and temporal reasoning. 
Video-oriented MLLMs, such as VideoChat2, Video-LLaVA, LLaVA-Video, and the Qwen-VL series, demonstrate strong capabilities in video instruction following, temporal understanding, object localization, and long multimodal context modeling~\cite{li2024mvbenchcomprehensivemultimodalvideo,lin2024videollavalearningunitedvisual,zhang2025llavavideovideoinstructiontuning,wang2024qwen2vlenhancingvisionlanguagemodels,bai2025qwen25vltechnicalreport,bai2025qwen3vltechnicalreport}.

However, standard video MLLM pipelines often rely on sparse frame sampling and holistic visual representations, which can easily omit critical action details~\cite{liu2024timecraft,liu2025commonsense,liu2026few}. To address this limitation, our work explicitly extracts action-centric cues that are highly indicative of the video's narrative impact. Supported by this structured information, the MLLM is then guided to effectively reason about the latent viewer sentiment.

\subsection{Structured and Grounded Video Evidence}
Recent works have explored structured video semantics and spatiotemporal action modeling to make actions, object interactions, and temporal event dynamics explicit~\cite{chen2025csta,huang}. 
Egocentric Action Scene Graphs represent long-form egocentric videos as temporally evolving action graphs with localized object instances~\cite{rodin2023actionscenegraphslongform}, while Video-LLM methods leverage spatio-temporal scene graphs or logical entailment structures for compositional reasoning and self-training~\cite{qiu2025stepenhancingvideollmscompositional,liu2025commonsense}.
Lightweight temporal-triplet methods further convert video semantics into ordered subject--predicate--object units~\cite{10.1145/3746252.3760857,chen2024mecd,chen2025mecd+}.

Our method shares the motivation of explicit scene modeling but differs in its selective focus and information density. Existing scene-graph methods aim for comprehensive scene reconstruction, building dense graphs that treat background elements, static objects, and actions indiscriminately. 
In contrast, our goal is to infer the latent emotion behind the narrative, so we bypass this dense and noisy representation. Our structured event representation selectively prioritizes the extraction of key action dynamics, the construction of their temporal evolution, and the reasoning over the latent emotional cues behind them.

\section{Method}
\begin{figure}[t]
  \centering
  \includegraphics[width=\textwidth,trim={0 1050 930 20},clip]{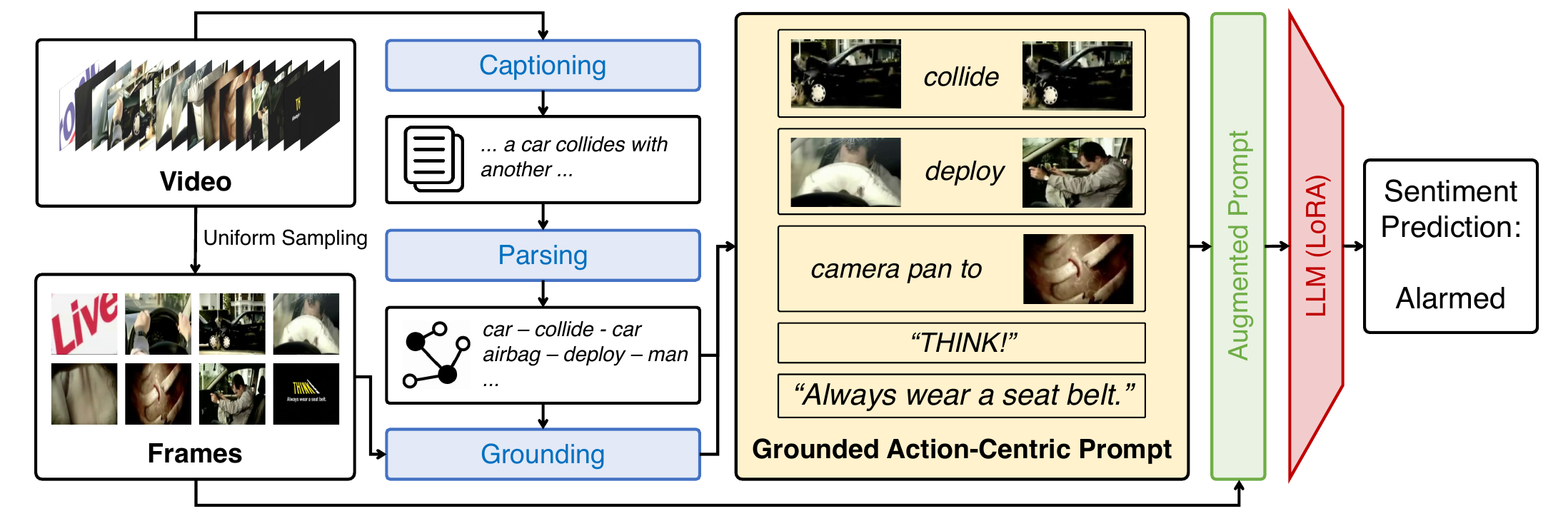}
  \caption{Overview of our grounded action-centric evidence augmentation framework.}
  \label{fig:model}
\end{figure}

To bridge the gap between observable advertisement cues and viewers' latent affective responses, we propose a grounded action-centric evidence augmentation framework. 
Given an input video, our method augments the standard sparse-frame MLLM input with temporally ordered subject--verb--object triplets, auxiliary visible textual cues, and localized visual evidence of subject and object entities, as illustrated in Figure~\ref{fig:model}. 
The augmented evidence is provided as structured multimodal context to the MLLM, leaving the backbone architecture unchanged.

Formally, given an advertisement video $V$, the MLLM receives sparse frames $X=\{x_k\}_{k=1}^{K}$, as in the baseline. 
For offline semantic extraction, we additionally sample dense frames $D=\{d_m\}_{m=1}^{M}$, where $M>K$. 
Dense frames are used only to extract semantic cues, while all grounded visual entity crops are obtained from $X$; thus, the augmented input does not introduce visual evidence beyond the baseline sparse input.
Our framework first extracts action triplets and visible textual cues from $D$, then grounds subject and object entities in $X$, and finally organizes the evidence as structured multimodal context for the MLLM.

\subsection{Action-Centric Semantic Extraction}
\label{sec:action_semantic_extraction}
Given the dense frame set $D$, we first use a frozen MLLM captioner $f_{\mathrm{cap}}$ with an action-focused instruction $I_{\mathrm{cap}}$ to generate a chronological action-centric caption $c=f_{\mathrm{cap}}(D;I_{\mathrm{cap}})$.
The instruction asks the model to describe only clearly visible evidence, with emphasis on actions and behaviors, i.e., who does what. 
We generate a free-form caption before triplet extraction because captioning is a stable MLLM capability and preserves the temporal flow of events.

We then use a frozen LLM parser $f_{\mathrm{parse}}$ with instruction $I_{\mathrm{parse}}$ to extract action triplets and auxiliary visible textual cues, $(\mathcal{E},\mathcal{T})=f_{\mathrm{parse}}(c;I_{\mathrm{parse}})$, where
\begin{equation}
\mathcal{E}=(e_1,\ldots,e_N),\quad
e_i=(s_i,v_i,o_i),\quad
\mathcal{T}=(t_1,\ldots,t_J).
\end{equation}
Here, $s_i$, $v_i$, and $o_i$ denote the subject, verb, and object of the $i$-th action triplet, and $o_i=\bot$ when no clear target entity is visible. 
We also retain visible textual cues, since slogans, product names, and promotional messages may provide affect-relevant cues in advertisements.

\subsection{Visual Entity Grounding}
\label{sec:visual_entity_grounding}
Since triplet entities are symbolic, we ground subject and object entities in the sparse frame set $X$ whenever possible, while keeping verbs as text.
For each triplet $e_i=(s_i,v_i,o_i)$, a frozen MLLM frame selector $f_{\mathrm{sel}}$ selects the sparse frame that best supports the event, producing $\tau_i=f_{\mathrm{sel}}(e_i,X;I_{\mathrm{sel}})$ with $\tau_i\in\{1,\ldots,K\}\cup\{\bot\}$.
Here, $\tau_i=\bot$ indicates that no sparse frame provides useful visual evidence.
This step aligns entity localization with the corresponding action event and reduces noisy grounding on irrelevant frames.

For each triplet, we ground the subject and, when $o_i\neq\bot$, the object entity. 
For each entity $r$ to be grounded, if $\tau_i\neq\bot$, we apply an open-set grounding model $f_{\mathrm{grd}}$ to the selected frame using $r$ as the query and keep the highest-confidence valid box $\hat{b}_i^r=\operatorname{TopValid}(f_{\mathrm{grd}}(x_{\tau_i},r))$. 
If $\tau_i=\bot$ or no valid box remains, we set $\hat{b}_i^r=\bot$.
The grounded representation is then defined as
\begin{equation}
g_i(r)=
\begin{cases}
\mathrm{Crop}(x_{\tau_i}, \hat{b}_i^r), & \text{if } \hat{b}_i^r \neq \bot,\\
r, & \text{otherwise}.
\end{cases}
\end{equation}

Thus, an entity is represented by a localized visual crop when reliable grounding is available and falls back to its textual name otherwise. 
To control computational overhead, we impose a crop budget $C$ over all successfully grounded entity instances in a video. 
When the number of valid crops exceeds $C$, we retain the top-$C$ crops ranked by grounding confidence and use textual fallbacks for the remaining entities.

\subsection{Grounded Action-Centric Evidence Augmentation}

We construct the final MLLM input by combining the original sparse video frames with the ordered grounded action-centric evidence. 
For each triplet $e_i=(s_i,v_i,o_i)$, we serialize the grounded action event as
\begin{equation}
z_i =
\begin{cases}
[\,g_i(s_i),\ v_i,\ g_i(o_i)\,], & \text{if } o_i \neq \bot,\\
[\,g_i(s_i),\ v_i\,], & \text{if } o_i = \bot.
\end{cases}
\end{equation}
Auxiliary visible textual cues are inserted with an explicit prefix, e.g., \texttt{On-screen text:}. 
Let $\mathcal{Z}=(z_1,\ldots,z_N)$ denote the serialized grounded action events. 
We merge $\mathcal{Z}$ and $\mathcal{T}$ according to their temporal order and construct the augmented MLLM input as
\begin{equation}
\mathcal{H}=\operatorname{Order}(\mathcal{Z},\mathcal{T}),\quad
\mathcal{P}_{\mathrm{ours}}=\operatorname{Concat}(X,\mathcal{H},q),
\end{equation}
where $q$ is the sentiment prediction question.

The MLLM is then asked to predict the viewer sentiment evoked by the advertisement video from the predefined emotion label space. 
For all comparisons, we use the same answer parsing protocol as the corresponding sparse-frame baseline. 
Implementation details are described in Section~\ref{sec:experimental_setup}.

\section{Experiments}
We evaluate our grounded action-centric evidence augmentation framework on video advertisement sentiment prediction, with main experiments on Pitts and transfer evaluations on AdsQA and an emotion-focused TVQA subset. 
We further conduct ablation, qualitative, and attention analyses to examine the contribution of structured action evidence and grounded visual entity crops.

\subsection{Experimental Setup}
\label{sec:experimental_setup}

\paragraph{Datasets and metrics.}
We conduct main experiments on 2,034 successfully downloaded videos from the Pitts dataset~\cite{hussain2017automatic}, using a 60/40 train--test split with seed 42. 
Following the dataset annotations, we report Acc$_{\text{clean}}$ against the majority-vote clean label and Acc$_{\text{raw}}$ against any normalized raw annotator label.
To evaluate generalization, we further test on AdsQA~\cite{long2025adsqa} and TVQA~\cite{lei2019tvqalocalizedcompositionalvideo}. 
For AdsQA, we use 998 emotion-oriented QA pairs and normalize free-form answers into the Pitts raw-label space, reporting Acc$_{\text{raw}}$. 
For TVQA, we construct a 1,536-sample emotion-focused subset, split it 60/40 with seed 42, and report Acc$_{\text{clean}}$.

\paragraph{Implementation details.}
We evaluate our method with two MLLM backbones, Qwen2.5-VL-7B and Qwen3-VL-8B.
Both the baseline and our method use $K=8$ uniformly sampled sparse frames as the final video input. 
Our offline evidence extraction uses $M=64$ dense frames for action-centric captioning, followed by triplet parsing, sparse-frame entity selection, and open-set grounding with frozen Qwen3-VL-8B-Instruct, Qwen2.5-14B-Instruct, and GroundingDINO~\cite{liu2023grounding}. 
All grounded crops are extracted only from the same sparse frame set used by the baseline, so no additional visual frames are introduced to the final MLLM input.

All models are fine-tuned for 3 epochs with AdamW, learning rate \texttt{1e-4}, cosine scheduling, BF16 precision, batch size 1, and gradient accumulation of 16. 
Results are averaged over three seeds, 41, 42, and 43. 
For Qwen2.5-VL, we set the crop budget to $C=24$ for memory feasibility and additionally use training-only input-format adaptation samples to stabilize interleaved visual-entity evidence; no additional test-time information is introduced.

\subsection{Main Results}
\label{sec:main_results}

\begin{table}[t]
\centering
\caption{Main results on the Pitts dataset.}
\label{table:main_results}
\begin{tabular}{lcc}
\toprule
Method & Acc$_{\text{clean}}$ & Acc$_{\text{raw}}$ \\ 
\midrule
\multicolumn{3}{l}{\textbf{Previous Methods and Closed-Source Models}} \\ 
\midrule
SVM$^{\ast}$ & 32.8 & -- \\
GPT-4o & 21.3 & 55.5 \\
\midrule
\multicolumn{3}{l}{\textbf{Fine-Tuned Open-Source MLLMs}} \\ 
\midrule
Qwen2.5-VL Baseline$^{\dagger}$ & 35.5 & 66.7 \\
Qwen2.5-VL + Ours$^{\ddagger}$ & 39.6 & 73.0 \\
Qwen3-VL Baseline$^{\dagger}$ & 37.2 & 69.3 \\
Qwen3-VL + Ours & 40.4 & 72.7 \\
\bottomrule
\end{tabular}

\smallskip
$^{\ast}$ From~\cite{hussain2017automatic};
$^{\dagger}$ sparse-frame fine-tuning;
$^{\ddagger}$ with auxiliary samples.
\end{table}

Table~\ref{table:main_results} compares our method with prior methods, GPT-4o, and fine-tuned open-source MLLM baselines on Pitts.
Fine-tuned MLLM baselines substantially outperform zero-shot GPT-4o, highlighting the need for adaptation to the advertisement viewer-sentiment label space. 
Our grounded action-centric evidence augmentation consistently improves both backbones: Qwen3-VL improves from 37.2 to 40.4 in Acc$_{\text{clean}}$ and from 69.3 to 72.7 in Acc$_{\text{raw}}$, while Qwen2.5-VL improves from 35.5 to 39.6 and from 66.7 to 73.0, respectively. 
These gains show the benefit of explicit action-event organization and grounded visual participants.

\subsection{Ablation Study}
\label{sec:ablation}
We ablate the components of our action-centric evidence using Qwen3-VL, comparing captions, text-only structured evidence $(\mathcal{E},\mathcal{T})$, and grounded evidence $\mathcal{H}$.
Qwen3-VL is used because it does not require auxiliary input-format adaptation, making the comparison cleaner.
Table~\ref{table:ablation_results} shows a consistent progression from sparse frames to captions, structured triplets, and grounded crops, confirming the complementary benefit of action structure and localized visual evidence.

\begin{table}[t]
\centering
\small
\caption{Ablation study of grounded action-centric evidence components.}
\label{table:ablation_results}
\begin{tabular}{lccc cc}
\toprule
Method & Cap. & Trip. & Crop & Acc$_{\text{clean}}$ & Acc$_{\text{raw}}$ \\
\midrule
Baseline (8 Frames) &  &  &  & 37.2 & 69.3 \\
+ Action-Centric Caption & \checkmark &  &  & 39.1 & 70.4 \\
+ Structured Text-Only Evidence &  & \checkmark &  & 39.5 & 71.8 \\
+ Grounded Visual Entity Crops (Ours) &  & \checkmark & \checkmark & 40.4 & 72.7 \\
\bottomrule
\end{tabular}

\smallskip
\footnotesize
Cap.: text video caption; Trip.: text-only triplets; Crop: grounded visual entity crops.
\end{table}

\subsection{Cross-Dataset and Cross-Task Transfer}
\label{sec:transfer}

We evaluate generalization by directly testing Pitts-trained checkpoints on the AdsQA emotion subset, and test transferability by fine-tuning and evaluating on the emotion-focused TVQA subset.
Tables~\ref{table:adsqa_results} and~\ref{table:tvqa_results} show consistent gains across both settings.
On AdsQA, our method improves Acc$_{\text{raw}}$ by 5.3 points for Qwen2.5-VL and 3.5 points for Qwen3-VL.
On TVQA, it improves Acc$_{\text{clean}}$ by 3.4 and 1.6 points, respectively, suggesting that grounded action-centric evidence can transfer beyond the Pitts test split and benefit broader emotion-related video reasoning tasks.

\begin{table}[t]
\centering

\begin{minipage}{0.45\linewidth}
\centering
\caption{Generalization on AdsQA.}
\label{table:adsqa_results}
\begin{tabular}{lc}
\toprule
Method & Acc$_{\text{raw}}$ \\
\midrule
Qwen2.5-VL Baseline & 44.7 \\
Qwen2.5-VL + Ours & 50.0 \\
Qwen3-VL Baseline & 47.2 \\
Qwen3-VL + Ours & 50.7 \\
\bottomrule
\end{tabular}
\end{minipage}
\hfill
\begin{minipage}{0.45\linewidth}
\centering
\caption{Transfer on TVQA.}
\label{table:tvqa_results}
\begin{tabular}{lc}
\toprule
Method & Acc$_{\text{clean}}$ \\
\midrule
Qwen2.5-VL Baseline & 74.3 \\
Qwen2.5-VL + Ours & 77.7 \\
Qwen3-VL Baseline & 76.8 \\
Qwen3-VL + Ours & 78.4 \\
\bottomrule
\end{tabular}
\end{minipage}
\end{table}

\subsection{Inference Budget Analysis}
We examine the downstream inference cost of our method on Pitts with Qwen3-VL by varying only the inference-time visual budget while keeping checkpoints fixed.
FLOPs exclude offline evidence extraction and are normalized by the original full-budget sparse-frame baseline.
Table~\ref{table:inference_budget} reports representative results.

\begin{table}[t]
\centering
\small
\setlength{\tabcolsep}{4pt}
\caption{Inference budget analysis on Pitts with Qwen3-VL.}
\label{table:inference_budget}
\begin{tabular}{lccccccc}
\toprule
Method & $K$ & VPix & $C$ & CPix & Acc$_{\text{clean}}$ & Acc$_{\text{raw}}$ & FLOPs$_{\text{norm}}$ \\
\midrule
Baseline (full) & 8 & full & 0 & -- & 37.2 & 69.3 & 1.00 \\
Baseline (matched) & 6 & 48 & 0 & -- & 34.9 & 66.1 & 0.47 \\
Ours (balanced) & 6 & 48 & 16 & 32 & 40.6 & 72.7 & 0.94 \\
Ours (efficient) & 6 & 48 & 8 & 32 & 40.4 & 73.1 & 0.79 \\
\bottomrule
\end{tabular}

\smallskip
\footnotesize
VPix/CPix: maximum visual patches per video frame/crop.
\end{table}

Reducing the video budget alone degrades the matched baseline, while our method surpasses the full baseline with lower FLOPs.
With $K=6$, VPix$=48$, and $C=8$, our method achieves 40.4 Acc$_{\text{clean}}$ and 73.1 Acc$_{\text{raw}}$ using only 0.79$\times$ FLOPs, suggesting that our gains come from more effective visual-budget allocation rather than simply using more visual tokens.

\subsection{Qualitative and Attention Analysis}
\label{sec:qualitative_attention}
We analyze how grounded action-centric evidence supports viewer sentiment prediction through a qualitative example and attention-based diagnostics.

\paragraph{Qualitative example.}
\begin{figure}[t]
  \centering
  \includegraphics[width=\textwidth,trim={0 470 230 30},clip]{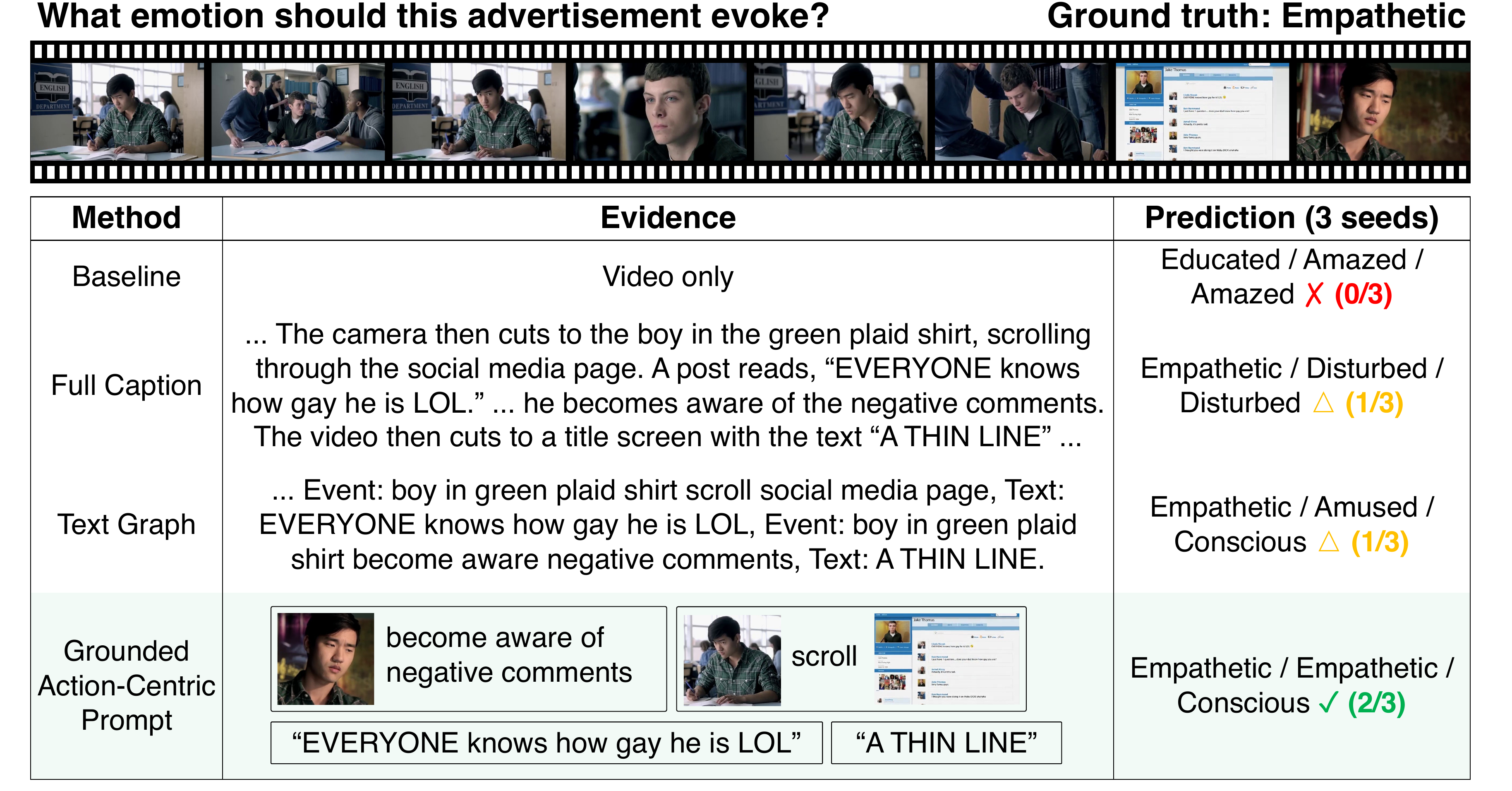}
  \caption{Qualitative example of an online-bullying advertisement. Our grounded action-centric evidence makes the victim-centered narrative explicit and helps infer the intended ``Empathetic'' viewer response.}
  \label{fig:ex2}
\end{figure}

\begin{figure}[t]
  \centering
  \includegraphics[width=\textwidth]{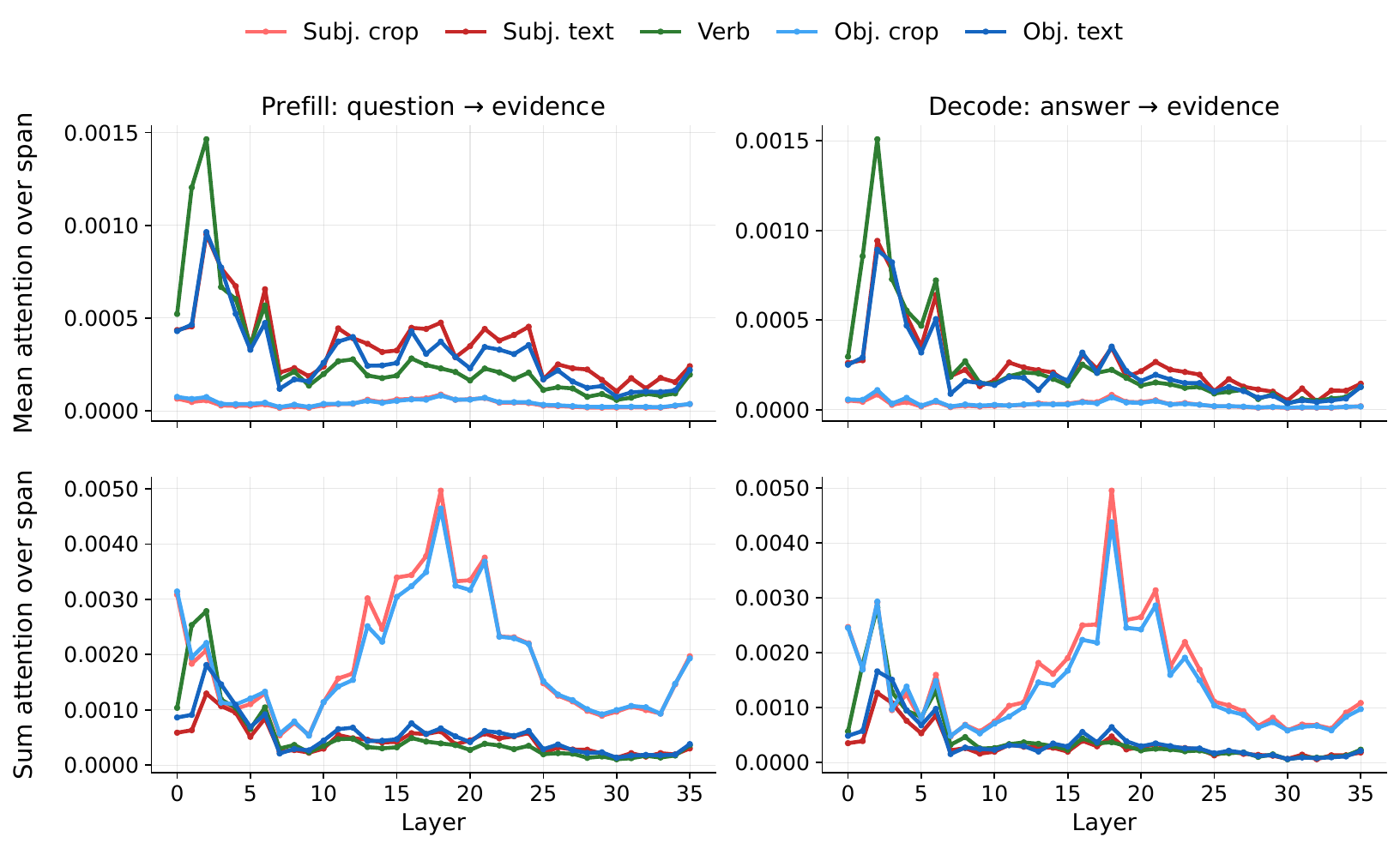}
\caption{
Layer-wise attention to grounded action-centric evidence components.
Columns show prefill and decoding attention; rows report mean and sum attention.
}
  \label{fig:attention_analysis}
\end{figure}

\begin{figure}[t]
  \centering
  \includegraphics[width=\linewidth]{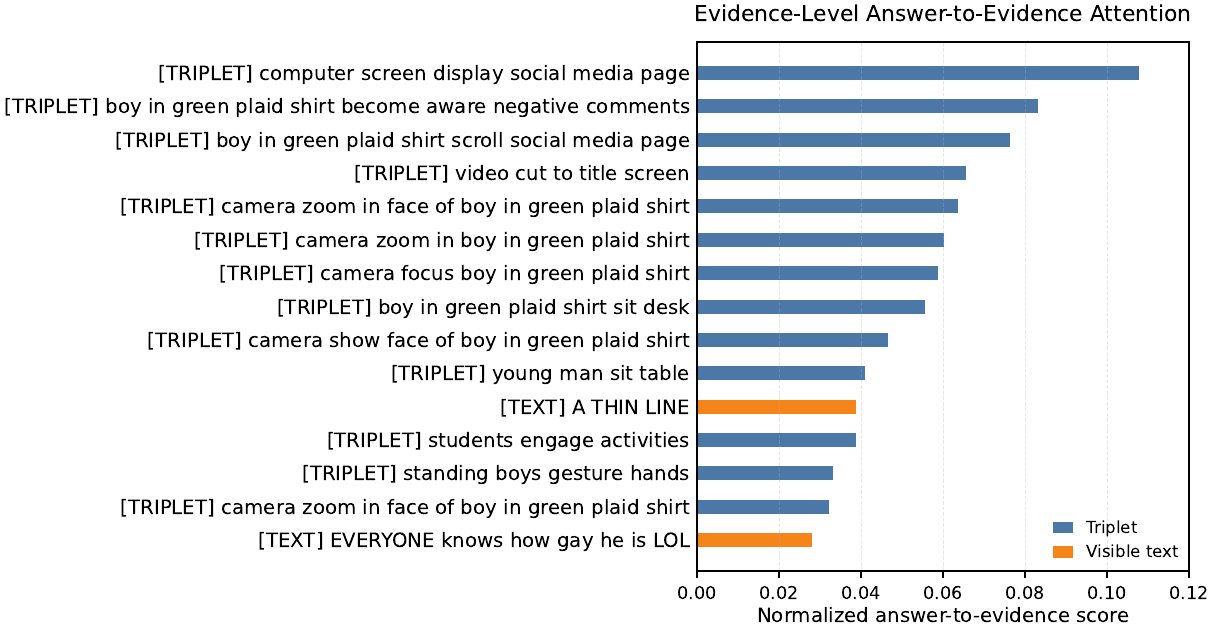}
  \caption{Evidence-level answer-to-evidence attention for the example in Fig.~\ref{fig:ex2}.}
  \label{fig:case_attention}
\end{figure}

Figure~\ref{fig:ex2} shows an online-bullying advertisement with the ground-truth viewer response ``Empathetic''.
The sparse-frame baseline fails to infer the intended viewer response, while caption and text-triplet evidence partially recover the relevant social-media bullying context. 
Our input evidence explicitly organizes the events ``scroll social media page'' and ``become aware of negative comments'', grounds the involved visual entities, and preserves visible textual cues such as ``EVERYONE knows how gay he is LOL'' and ``A THIN LINE''. 
These cues make the victim-centered narrative more explicit, leading to correct predictions in two out of three seeds.

\paragraph{Component-level attention.}
To inspect how the model uses the inserted grounded action-centric evidence, we analyze attention to component spans within the grounded evidence sequence $\mathcal{H}$. 
We group grounded-triplet tokens into visual subjects, textual subjects, verbs, visual objects, and textual objects, and compute attention from question tokens during prefill and answer tokens during decoding.
Let $A^{(\ell)}$ denote the attention matrix at layer $\ell$ after averaging over attention heads. 
Given a query-token set $Q$ and a component span $S$, we compute mean and sum attention over the span as
\begin{equation}
a_{\mathrm{mean}}^{(\ell)}(Q,S)
=
\frac{1}{|Q|\,|S|}
\sum_{u \in Q}\sum_{k \in S} A^{(\ell)}_{u,k},
\quad
a_{\mathrm{sum}}^{(\ell)}(Q,S)
=
\frac{1}{|Q|}
\sum_{u \in Q}\sum_{k \in S} A^{(\ell)}_{u,k}.
\end{equation}

Scores are averaged over component instances within each video and then across test videos.
Figure~\ref{fig:attention_analysis} shows that verbs receive high mean attention in shallow layers, while grounded crops receive substantial summed attention in middle layers, suggesting that the model uses both action structure and localized perceptual evidence.
Although diagnostic rather than causal, these patterns support our motivation that action triplets expose event structure while grounded crops provide usable visual details.

\paragraph{Evidence-level attention.}
For the example in Fig.~\ref{fig:ex2}, we further aggregate normalized answer-to-evidence attention from generated answer tokens to each inserted triplet or visible textual cue.
As shown in Fig.~\ref{fig:case_attention}, the highest-scoring evidence units form a coherent online-bullying narrative: a social-media page is displayed, the boy becomes aware of negative comments, and he scrolls through the page. 
On-screen text such as ``A THIN LINE'' and ``EVERYONE knows how gay he is LOL'' further complements this victim-centered narrative.

Overall, these analyses suggest that grounded action-centric evidence provides interpretable intermediate evidence for affective video reasoning.

\section{Conclusion}
We studied viewer sentiment prediction in video advertisements, where the goal is to infer latent audience response from observable video cues rather than recognize the emotions of visible subjects. 
We proposed a grounded action-centric evidence augmentation framework that represents videos with temporally ordered action triplets and visible textual cues, grounds subject and object entities as localized visual evidence, and organizes the resulting evidence as structured multimodal context for the MLLM without modifying the underlying architecture.
Experiments on Pitts show consistent improvements over sparse-frame MLLM baselines, while ablations confirm the complementary benefits of structured action evidence and grounded visual entity crops.
Cross-dataset evaluation on AdsQA and transfer experiments on an emotion-focused TVQA subset further suggest that grounded action-centric evidence augmentation provides transferable evidence for broader emotion-related video reasoning.
Overall, our results highlight the importance of integrating symbolic action-event structure with localized visual evidence for sentiment prediction in video advertisements.

\bibliographystyle{splncs04}
\bibliography{bibliography}

\end{document}